%% file: arxiv_HMR.tex
\documentclass[a4paper,11pt]{article}
\usepackage[top=4cm, bottom=4cm, left=3cm, right=3cm]{geometry}

\usepackage[utf8]{inputenc} 
\usepackage[T1]{fontenc}    
\usepackage{url}            
\usepackage{amsfonts}       
\usepackage{amssymb}
\usepackage{amsmath}
\usepackage{amsthm}
\usepackage{mathtools}
\usepackage{tikz}
\usepackage{subcaption}
\usepackage{bm}

\usepackage{mathrsfs} 
\usepackage[usenames,dvipsnames]{pstricks}
\usepackage{euler}
\usepackage{euscript}
\usepackage{graphicx}
\usepackage{charter}
\usepackage{mdframed}
\usepackage{enumerate, color}
\usepackage[round]{natbib}

\usepackage[hyperindex]{hyperref}
\usepackage{newfloat}

\usepackage{multirow}
\usepackage[nameinlink,capitalize]{cleveref}
\usepackage{floatrow}
\usepackage{graphicx}
\usepackage{subcaption}
\usepackage{booktabs}

\usepackage{cleveref}

\usepackage{algorithm, algorithmic}

\usepackage[font=small,skip=0pt]{caption}
\usepackage{thmtools, thm-restate}
\graphicspath{ {pictures/} }

\input{defs}

\title{\sffamily\huge\bf Hyperbolic Manifold Regression}

\author{ Gian Marconi Marconi$^{1}$ \and Lorenzo Rosasco$^{1,3,4}$ \and Carlo Ciliberto$^{2}$}

\begin{document}
    \maketitle

\begin{abstract}
    \input{sections/abstract.tex}
\end{abstract}

\section{Introduction}
\label{Instroduction}
\footnotetext[1]{Istituto Italiano di Tecnologia, Via Morego, 30, Genoa 16163, Italy.}\footnotetext[2]{Imperial College London, SW7 2BU London, United Kingdom}\footnotetext[3]{Massachusetts Institute of Technology, Cambridge, MA 02139, USA.}\footnotetext[4]{Universit\'a degli Studi di Genova, Genova, Italy.}

\input{sections/introduction.tex}

\section{Hyperbolic Representations}
	\label{sec:hyper}
\input{sections/hyperbolic_data_representations.tex}

\section{Manifold Valued Prediction in Hyperbolic Space}
	\label{sec:method}
\input{sections/manifold_valued_regression.tex}

\section{Experiments}
	\label{sec:experiments}
\input{sections/applications.tex}

\section{Conclusion}
	\label{sec:conclusions}
\input{sections/conclusions.tex}

\section*{Acknowledgments}

We thank Maximilian Nickel for his invaluable support and feedback throughout this project. Without him, this paper would have not been possible.

\small{This material is based upon work supported by the Center for Brains, Minds and Machines (CBMM), funded by NSF STC award CCF-1231216, and the Italian Institute of Technology. We gratefully acknowledge the support of NVIDIA Corporation for the donation of the Titan Xp GPUs and the Tesla k40 GPU used for this research.
    This work has been carried out at the Machine Learning Genoa (MaLGa) center, Università di Genoa (IT)
    L. R. acknowledges the financial support of the European Research Council (grant SLING 819789), the AFOSR projects FA9550-17-1-0390  and BAA-AFRL-AFOSR-2016-0007 (European Office of Aerospace Research and Development), and the EU H2020-MSCA-RISE project NoMADS - DLV-777826.
}

\bibliographystyle{abbrvnat}
\bibliography{../HSP}

\end{document}

%% file: defs.tex
\providecommand{\nor}[1]{\left\|{#1}\right\|}

\newcommand\manifold[1]{\mathcal{#1}}

\newcommand{\R}{\mathbb{R}}

\DeclareMathOperator*{\argmin}{arg\,min}
\renewcommand\vec[1]{\bm{#1}}
\newcommand\vx{\vec{x}}
\newcommand\vy{\vec{y}}
\newcommand\vv{\vec{v}}
\newcommand\vu{\vec{u}}
\newcommand\vz{\vec{z}}
\newcommand{\grad}{\operatorname{grad}}
\newcommand{\tansp}{\mathcal{T}}
\newcommand{\Lor}{\manifold{L}}
\newcommand\arcosh{\operatorname{acosh}}
\newcommand\ldot[1]{\langle #1 \rangle_{\Lor}}
\newcommand{\lnorm}[1]{\| #1 \|_{\Lor}}

\newcommand{\X}{{\mathcal X}}
\newcommand{\Y}{{\mathcal Y}}

\newcommand{\risk}{{\mathcal E}}

\newcommand{\fhsp}{{f_{hsp}}}

\newcommand{\mbf}[1]{\mathbf{#1}}
\newcommand{\msf}[1]{\mathsf{#1}}

\providecommand{\nor}[1]{\bigl\|{#1}\bigr\|}

\renewcommand{\paragraph}[1]{{\bf #1.}}

\declaretheorem[name=Theorem,refname=Thm.]{theorem}

\declaretheorem[name=Proposition,refname=Prop.,sibling=theorem]{proposition}

\declaretheorem[name=Corollary,refname=Cor.,sibling=theorem]{corollary}

\declaretheorem[name=Assumption,refname=Asm.]{assumption}
\crefname{assumption}{assumption}{assumptions}

\crefname{equation}{}{Eqs.}
\Crefname{equation}{Eq.}{Eqs.}
\crefname{figure}{Fig.}{Figs.}


%% file: sections/abstract.tex




Geometric representation learning has recently shown great promise in several machine learning settings, ranging from relational learning to language processing and generative models. 
In this work, we consider the problem of performing manifold-valued regression onto an hyperbolic space as an intermediate component for a number of relevant machine learning applications. In particular, by formulating the problem of predicting nodes of a tree as a manifold regression task in the hyperbolic space, we propose a novel perspective on two challenging tasks: $1$) hierarchical classification via label embeddings and $2$) taxonomy extension of hyperbolic representations. To address the regression problem we consider previous methods as well as proposing two novel approaches that are computationally more advantageous: a parametric deep learning model that is informed by
the geodesics of the target space and a non-parametric kernel-method for which we also prove
excess risk bounds. Our experiments show that the strategy of leveraging the hyperbolic geometry is promising. In particular, in the taxonomy expansion setting, we find that the hyperbolic-based estimators significantly outperform methods performing regression in the ambient Euclidean space.


%% file: sections/introduction.tex
Representation learning is a key paradigm in machine learning and artificial
intelligence. It has enabled important breakthroughs in
computer vision~\citep{krizhevsky2012imagenet,he2016deep}
natural language processing~\citep{DBLP:journals/corr/MikolovSCCD13,bojanowski2016enriching,joulin2016bag},
relational learning~\citep{nickel2011threeway,perozzi2014deepwalk},
generative modeling~\citep{kingma2013auto,radford2015unsupervised},
and many other areas \citep{bengio2013representation,lecun2015deep}.
Its objective is typically to infer latent feature representations of objects
(e.g., images, words, entities, concepts) such that their similarity or distance
in the representation space captures their \emph{semantic} similarity.
For this purpose, the geometry of the representation space has recently received
increased
attention~\citep{wilson2014spherical,falorsi2018explorations,davidson2018hyperspherical,xu2018spherical}.
Here, we focus on Riemannian representation spaces and in particular on
hyperbolic geometry. \cite{Nickel2017a} introduced Poincar\'{e} embeddings to
infer hierarchical representations of symbolic data, which led to substantial gains in representational efficiency and
generalization performance. Hyperbolic representations have since been
extended to other manifolds~\citep{Nickel2018,DeSa2018}, word embeddings~\citep{tifrea2018poincar,le2019inferring}, recommender
systems~\citep{chamberlain2019scalable}, and image
embeddings~\citep{khrulkov2019hyperbolic}.

However, it is yet an open problem how to efficiently integrate hyperbolic
representations with standard machine learning methods which often make a
Euclidean or vector space assumption. The work of \cite{Ganea2018} establishes some fundamental
steps in this direction by proposing a generalization of
fully connected neural network layers from Euclidean space to hyperbolic space.
However most of the experiments shown were from hyperbolic to Euclidean space 
using recurrent models.
In this paper we focus on the
task of learning manifold-valued functions from Euclidean on to hyperbolic space
that allows us to leverage its hierarchical structure for supervised learning.
For this purpose, we propose two novel approaches: a deep learning model trained with a geodesic-based loss to learn hyperbolic-valued functions and a non-parametric kernel-based
model for which we provide a theoretical analysis. 

We illustrate the effectiveness of this strategy on two challenging tasks, i.e.,
hierarchical classification via label embeddings and taxonomy expansion by
predicting concept embeddings from text.
For standard classification tasks, label embeddings have shown great promise as
they allow to scale supervised learning methods to datasets with massive label
spaces~\citep{chollet2016information,veit2018separating}.
By embedding labels in hyperbolic space according to their natural hierarchical
structure (e.g, the underlying WordNet taxonomy of ImageNet
labels) we are then able to combine the benefits of hierarchical classification
with the scalability of label embeddings.
%
%
Moreover, the continuous nature of hyperbolic space allows the model to \emph{invent}
new concepts by predicting their placement in a pre-embedded base taxonomy. We
exploit this property for a novel task which we refer to as {\em taxonomy expansion}:
Given an embedded taxonomy $\mathcal{T}$, we infer the placement of unknown novel concepts by predicting their
features onto the embedding. In contrast to hierarchical classification, the
predicted embeddings are here full members of the taxonomy, i.e., they can
themselves act as parents of other points.
%
For both tasks, we show empirically that the proposed strategy can often lead to more effective estimators than its Eucliean counterpart. These findings support the thesis of this work that leveraging the hyperbolic geometry can be advantageous for several machine learning settings. Additionally, we observe that the hyperbolic-based estimators introduced in this work achieve comparable performance to the previously proposed hyperbolic neural networks \citep{Ganea2018}. This suggests that, in practice, it is not necessary to work with hyperbolic layers as long as the training procedure exploits the geodesic as an error measure. This is advantageous from the computational perspective, since we found our proposed approaches to be generally significantly easier to train in practice.


The remainder of this paper is organized as follows: In \Cref{sec:hyper} we briefly
review hyperbolic embeddings and related concepts such as Riemannian
optimization. In \Cref{sec:method}, we introduce our proposed methods and prove
excess risk bounds for the kernel-based method. In \Cref{sec:experiments} we
evaluate our methods on the tasks of hierarchical classification and taxonomy
expansion.


%% file: sections/hyperbolic_data_representations.tex





Hyperbolic space is the unique, complete, simply con-
nected Riemannian manifold with constant negative
sectional curvature. There exist multiple equivalent
models for hyperbolic space. To estimate the embed-
dings using stochastic optimization we will employ the
Lorentz model due to its numerical advantages. For
analysis, we will map embeddings into the Poincaré
disk which provides an intuitive visualization of hyper-
bolic embeddings. This can be easily done because the
two models are isometric~\cite{Nickel2018}. We
review both manifolds in the following.


\paragraph{Lorentz Model} Let \(\vu\), \(\vv \in \R^{n+1}\) and let $\ldot{\vu,
  \vv} = -u_0 v_0 + \sum_{i=1}^n u_n v_n$ denote the \emph{Lorentzian scalar
  product}. The Lorentz model of \(n\)-dimensional hyperbolic space is then
defined as the Riemannian manifold \(\Lor^n = (\manifold{H}^n, g_\Lor)\),
where
\begin{equation}
  \manifold{H}^n = \{\vu \in \R^{n+1} : \ldot{\vu, \vu} = -1, x_0 > 0\},
\end{equation}
denotes the upper sheet of a two-sheeted \(n\)-dimensional hyperboloid and where
$g_\Lor(\vu) = \operatorname{diag}([-1, 1, \ldots, 1])$ is the associated metric tensor.
Furthermore, the distance on $\Lor$ is defined as
\begin{equation}
  d_\Lor(\vu, \vv) = \arcosh(-\ldot{\vu, \vv}). \label{eq:ldist}
\end{equation}
An advantage of the Lorentz model is that its exponential map has as simple,
closed-form expression. As showed by \cite{Nickel2018}, this allows us to perform
Riemannian optimization efficiently and with increased numerical stability.
In particular, let $\vu \in \Lor^n$ and let $\vz \in \tansp_{\vu} \Lor^n$ denote
a point in the associated tangent space. The exponential map
\({\exp_{\vu} : \tansp_{\vu} \Lor^n \to \Lor^n}\) is then
defined as
\begin{equation}
  \exp_{\vu}(\vz) = \cosh(\lnorm{\vz})\vu + \sinh(\lnorm{\vz})\frac{\vz}{\lnorm{\vz}}.
  \label{eq:expm}
\end{equation}

\paragraph{Poincaré ball} The Poincaré ball model is the Riemannian manifold
\(\manifold{P}^n = (\manifold{B}^n, g_p)\), where \({\manifold{B}^n = \{\vu \in \R^n : \|\vu\| < 1\}}\)
is the \emph{open} \(n\)-dimensional unit ball and where $g_p(\vu) = 4 / ({1 - \|\vu\|^2})^2$ is the associated metric tensor. The distance function on \(\manifold{P}\) is defined as
\begin{equation}
d_p(\vu, \vv) = \arcosh \left(1 + 2 \frac{\|\vu - \vv\|^2}{(1 - \|\vu\|^2)(1 - \|\vv\|^2)} \right).
\label{eq:pdist}
\end{equation}

An advantage of the Poincaré ball is that it provides an intuitive model of
hyperbolic space which is well suited for analysis and visualization of the
embeddings. It can be seen from \Cref{eq:pdist}, that the distance within the
Poincaré ball changes smoothly with respect to the norm of \(\vu\) and \(\vv\).
This locality property of the distance is key for representing hierarchies
efficiently~\citep{hamann_2018}. For instance, by placing the root node of a tree at the origin of
\(\manifold{B}^n\), it would have relatively small distance to all other nodes,
as its norm is zero. On the other hand, leaf nodes can be placed close to the
boundary of the ball, as the distance between points grows quickly with a norm
close to one.

\paragraph{Hyperbolic embeddings}
\label{par:hyperbolic_embeddings}
We consider supervised datasets $\{x_i, c_i\}_{i=1}^m \in \mathcal{X} \times
\mathcal{C}$ where class labels $c_{i}$ can be organized according to a
taxonomy or class hierarchy $\mathcal{T} = (\mathcal{C}, \mathcal{E})$. Edges $(i,j) \in \mathcal{E}$ indicate that $c_i$ \emph{is-a} $c_j$. To compute hyperbolic embeddings of
all $c_{i}$ that capture these hierarchical relationships of $\mathcal{T}$, we
follow the works of \cite{Nickel2017a, Nickel2018} and infer the embedding from pairwise
similarities. In particular, let \(\gamma: \mathcal{C} \times \mathcal{C} \to \R_+ \)
be the similarity function such that
\begin{equation}
  \gamma(c_{i}, c_{j}) = \begin{cases}1, & \text{if $c_{i}$, $c_{j}$ are adjacent in $clos(\mathcal{T})$} \\
	0, & \text{otherwise}
	\end{cases}
\end{equation}
where $clos(\mathcal{T})$ is the transitive closure of $\mathcal{T}$.
Furthermore, let \(\mathcal{N}(i,j) = \{\ell : \gamma(i, \ell) <
\gamma(i,j)\} \cup \{j\}\) denote the set of concepts that are \emph{less} similar to \(c_i\)
then \(c_j\) (including \(c_j\)) and
let $\phi(i, j) = \argmin_{k \in \mathcal{N}(i, j)} d(\vu_i, \vu_k)$
denote the nearest neighbor of \(c_i\) in the set \(\mathcal{N}(i,j)\). We then learn
embeddings \(\Theta = \{\vu\}_{i=1}^m\) by optimizing
\begin{equation}
  \min_{\Theta} -\sum_{i,j} \log \Pr(\phi(i,j) = j\ |\ \Theta)
\end{equation}
with
\begin{equation}
  \Pr(\phi(i,j) = j\ |\ \Theta)= \frac{e^{d(\vu_i, \vu_j)}}{\sum_{k \in \mathcal{N}(i,j)} e^{d(\vu_i, \vu_k)}}.
  \label{eq:loss}
\end{equation}
\Cref{eq:loss} can be interpreted as a ranking loss that aims to extract latent hierarchical structures
from \(\mathcal{C}\). For computational efficiency, we follow \cite{jean2014using} and randomly subsample \(\mathcal{N}(i,j)\) on large datasets.
To infer the embeddings $\theta$ we then minimize \Cref{eq:loss} using Riemannian
SGD~\citep{optimization/bonnabel2013stochastic}. In RSGD, updates to the
parameters \(\theta\) are computed via
\begin{equation}
  \theta_{t+1} = \exp_{\theta_t}\bigg(-\eta \sum\limits_{j \in B}\grad_{\manifold{L}} f_j(\theta_t)\bigg), \label{eq:RSGD_iteration}
\end{equation}
where \(\grad_{\manifold{L}} f (\theta_t) \in \tansp_\theta \Lor\) denotes the
\emph{Riemannian gradient}, \(\eta\) denotes the learning rate, and $B = [j_1, \ldots, j_B] $ is a set of random uniformly sampled indexes.

\begin{figure*}[!h]
    \includegraphics[width=0.885\textwidth]{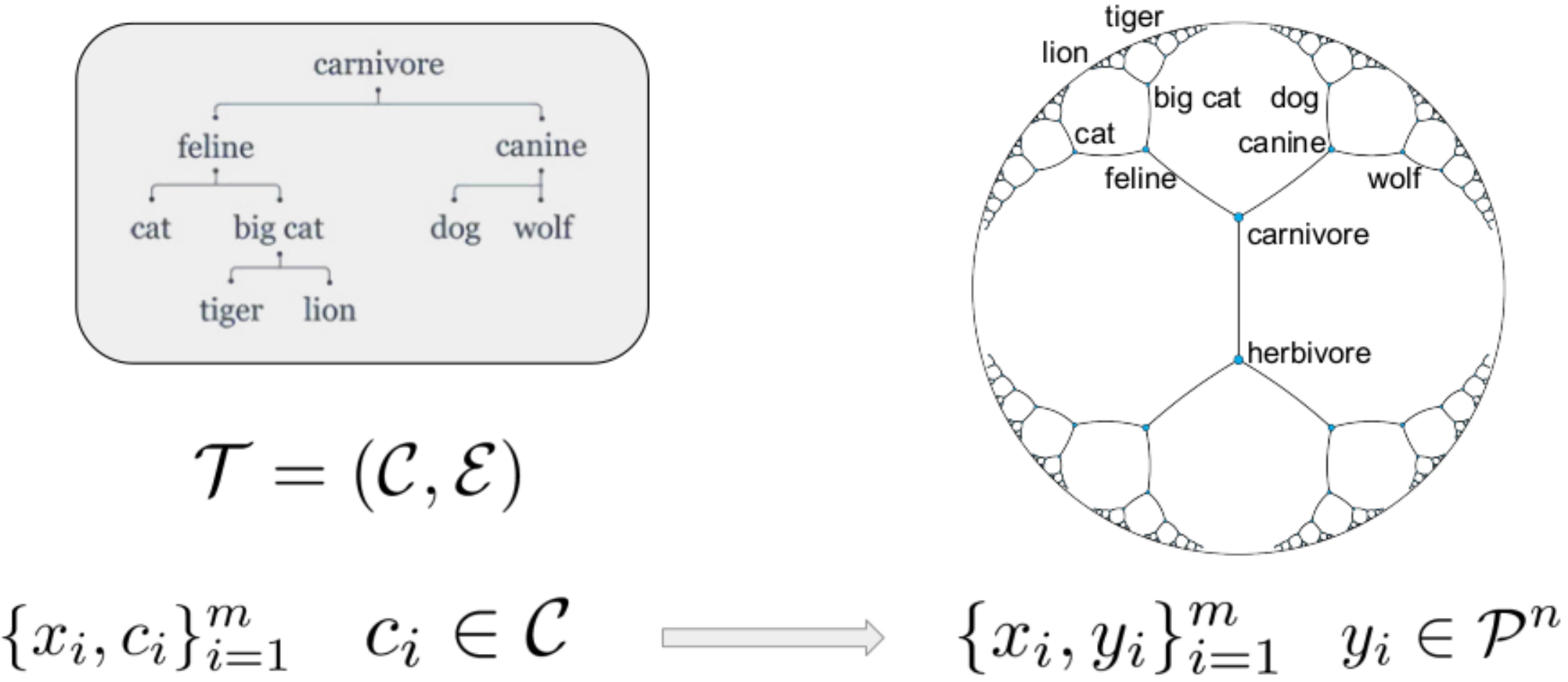}
    \caption{Hyperbolic representation of a hierarchical dataset}
    \label{fig:hdr}
\end{figure*}

By computing hyperbolic embeddings of $\mathcal{T}$, we have then recast the
learning problem from a discrete tree $\mathcal{D} = \{x_i, c_i\}_{i=1}^m, c_i \in \mathcal{C}$ to its embedding in a continuous manifold $\mathcal{D}^{e} =\{x_i, y_i\}_{i=1}^m$ with $y_i \in \mathcal{L}$. This
allows us to apply manifold regression techniques as discussed in the
following. This idea is depicted in~\cref{fig:hdr}.

%% file: sections/manifold_valued_regression.tex
We study the problem of learning $f:\X\to\Y\subset\Lor^n$ a map taking values in the hyperbolic space, often referred to as {\em manifold regression} \citep{steinke2009non,steinke2010nonparametric}. We assume for simplicity that $\X \subset \R^d$ and $\Y\subset\Lor^n$ are compact subsets. In particular, we assume a training dataset $\{\vx_i,\vy_i\}_{i=1}^m$ of points independently sampled from a joint distribution $\rho$ on $\X\times\Y$ and aim to find an estimator for the minimizer of the expected risk
\begin{equation}\label{eq:expected-risk}
	\min_{f:\X\to\Y}~ \risk(f)\quad \risk(f) = \int d_\Lor(f(\vx), \vy)^2 ~ d\rho(\vx,\vy).
\end{equation}
Here we consider $\Lor^n$ as target space and $d_\Lor$ as loss function, but all results extend to $\manifold{P}$. \Cref{eq:expected-risk} is the natural generalization of standard vector-valued ridge regression (indeed  the geodesic of $\Y = \R^n$ is the Euclidean distance $d(\vx,\vy) = \nor{\vx - \vy}$). We tackle this problem proposing two novel approaches: one leveraging recent results on structured prediction and one using geodesic neural networks.

\paragraph{Structured Prediction} \cite{Rudi2018} proposed a new approach to address manifold regression problems. The authors adopted a perspective based on structured prediction and interpreted the target manifold $\Y$ as a ``structured'' output. While standard structured prediction studies settings where $\Y$ is a discrete (often finite) space~\citep{bakir2007predicting}, this extension allowed the authors to design a kernel-based approach for structured prediction for which they provided a theoretical analysis under suitable assumptions on the output space. We formulate the corresponding Hyperbolic Structured Prediction (HSP) estimator when applying this strategy to our problem (namely $\Y \subset \Lor^n$). In particular, we have $\fhsp:\X\to\Lor^n$ the function such that for any test point $\vx\in\X$
\begin{equation}\label{eq:hsp}
	\fhsp(\vx) = \argmin_{\vy\in\Lor^n} ~ \sum_{i=1}^m \alpha_i(\vx)~d_\Lor(\vy,\vy_i)^2,
\end{equation}
where the weights $\alpha(\vx) = (\alpha_1(\vx),\dots,\alpha_n(\vx))^\top\in\R^m$ are learned by solving a variant of kernel ridge regression: given $k:\X\times\X\to\R$ a reproducing kernel on the input space, we obtain
\begin{equation}\label{eq:alphas}
	\alpha(\vx) = (\mbf{K} + \lambda \mbf{I})^{-1} \mbf{v}(\vx),
\end{equation}
where $\mbf{K}\in\R^{m \times m}$ is the empirical kernel matrix and $\mbf{v}\in\R^m$ is the evaluation vector with entries with entries respectively $\mbf{K}_{ij} = k(\vx_i,\vx_j)$ and $\mbf{v}(\vx)_i = k(\vx,\vx_i)$ for $i,j\in\{1,\dots,m\}$. 

In line with most literature on structured prediction, the estimator in \cref{eq:hsp} requires solving an optimization problem at every test point. Hence, while this approach offers a significant advantage at training time (when learning the weights $\alpha$), it can lead to a more expensive operation at test time. To solve this problem in practice we resort to RSGD as defined in \cref{eq:RSGD_iteration}.

\cite{Rudi2018}, studied the generalization properties of estimators of the form of \cref{eq:hsp}. The authors proved that under suitable assumptions on the regularity of the output manifold, it was possible to give bounds on the excess risk in terms of the number of training examples available. The following theorem specializes this result to the case of $\fhsp$. A key role will be played by the $(s,2)$-Sobolev space $W^{s,2}$ of functions from $\Lor^n$ to $\R$, which generalizes the standard notion on Euclidean domains~\cite[see][]{hebey2000nonlinear}.

\begin{theorem}\label{thm:rates}
Let $(\vx_i,\vy_i)_{i=1}^m$ be sampled independently according to $\rho$ on $\X\times\Y$ with $\Y\subset\Lor^m$ compact sets. Let $\fhsp$ defined as in \cref{eq:hsp} with weights \cref{eq:alphas} learned with reproducing kernel $k:\X\times\X\to\R$ with reproducing kernel Hilbert space (RKHS) $\mathcal{F}$. If the map ${x\mapsto\int d_\Lor(\cdot,\vy)^2~d\rho(y|x)}$ belongs to $W^{s,2}(\Y)\otimes\mathcal{F}$ with $s>n/2$, then for any $\tau\in(0,1]$
\begin{equation}\label{eq:rates}
	\risk(\fhsp) - \inf_{f}\risk(f) ~~\leq~~ \nor{d_\Lor^2}_{s,2} \msf{q}\tau^2~\frac{1}{n^{1/4}},
\end{equation}
holds with probability at least $1-8e^{-\tau}$, where $\msf{q}$ is a constant not depending on $n,\tau$ or $\nor{d_\Lor}_{s,2}$. 
\end{theorem}

The result guarantees a learning rate of order $O(n^{-1/4})$. We comment on the assumptions and constants appearing in \cref{thm:rates}. First, we point out that, albeit the requirement $\int d_\Lor(\cdot,\vy)^2~d\rho(y|x)\in\mathcal{F}\otimes W^{s,2}(\Y)$ can seem overly abstract, it reduces to a standard assumption in statistical learning theory. Informally, it corresponds to a regularity assumption on the conditional mean embedding of the distribution $\rho(\cdot|x)$ (see the work of~\cite{song2013kernel} for more details), and can be interpreted as requiring the solution of \cref{eq:expected-risk} to belong to the hypotheses space associated to the kernel $k$. Second, we comment on the constant in \cref{eq:rates} that depending on the geodesic distance. In particular, we note that by Thm. 2 of~\cite{Rudi2018} the squared geodesic on any compact subset of $\Lor^n$ belongs to $W^{s,2}(\Y)$ for any $s\geq0$. Hence $\nor{d_\Lor^2}_{s,2}<+\infty$ also for any $s>n/2$, as required by \cref{thm:rates}.

\begin{proof}
The proof of \cref{thm:rates} is a specialization of Thm. 2 and 4 by~\cite{Rudi2018}. We recall a key assumption that is required to apply such results. 

\begin{assumption}\label{asm:manifold-regularity}
	$\mathcal{M}$ is a complete $n$-dimensional smooth connected Riemannian manifold,
	without boundary, with Ricci curvature bounded below and positive injectivity radius.
\end{assumption}

The assumption above imposes basic regularity conditions on the output manifold. A first implication is indeed that. 

\begin{proposition}[Thm. $2$ in \cite{Rudi2018}]\label{prop:geod-smooth}
	Let $\manifold{M}$ satisfy \cref{asm:manifold-regularity} and let $\mathcal{Y}\subset\manifold{M}$ is a compact geodesically convex subset of $\manifold{M}$. Then, the squared geodesic distance $d:\manifold{M}\times\manifold{M}\to\R$ is smooth on $\mathcal{Y}$. Moreover, by the proof of Thm.$1$ in the appendix of \textit{Manifold Structured Prediction}~\citep{Rudi2018}, we have $d^2\in W^{s,2}(\mathcal{Y})$ for any $s>n/2$. 
\end{proposition} 

Leveraging standard results from Riemannian geometry, we can guarantee that the manifolds considered in this paper satisfy the above requirements. For simplicity, we restrict on $\manifold{M}$ corresponding to an open bounded ball in either $\mathcal{P}^n$ or $\Lor^n$. In particular,
\begin{itemize}
	\item $\manifold{M}$ has sectional curvature constantly equal to $-1$. Hence the Ricci curvature is bounded from below since we are in a bounded ball in either $\mathcal{P}^n$ or $\Lor^n$. 
	\item The injectivity radius is positive (actually lower bounded by $1/(2\cdot 9^{2 + [n/2]})$ with $[n/2]$ the integer parts of $n/2$), see Main Theorem by~\cite{martin1989balls}.
\end{itemize}

We see that we are in the hypotheses of \cref{prop:geod-smooth}, from which we conclude the following.

\begin{corollary}\label{cor:geod-smooth}
	For any $s\geq 0$, the geodesic distance $d_\Lor$ (respectively $d_{\mathcal{P}}$) belongs to $W^{s,2}(\Y)$ for any compact subspace of $\Lor^n$ (respectively $\mathcal{P}^n$).  
\end{corollary}

This guarantees us that we are in the hypotheses of \cite[Thm. $4$]{Rudi2018}, from which \cref{thm:rates} follows. We note in particular that $d_\Lor^2$ takes the role of the loss function $\bigtriangleup$ in the original theorem. Which needs to be a so-called ``Structure Encoding Loss Function''. The latter is guaranteed by \cref{cor:geod-smooth} above. 
\end{proof}

\paragraph{Neural Network with Geodesic loss (NN-G)} As an alternative to the
non-parametric model $f_{hsp}$, we consider also a parametric method based on
deep neural networks. An important challenge when dealing with manifold
regression is how to design a suitable model for the estimator. While neural
networks of the form $g_\theta \colon \R^d \to \R^k$ (parametrized by some weights $\theta$) have proven to be powerful models for
regression and feature
representation~\citep{lecun2015deep,bengio2013representation,xiao2016learning,ngiam2011multimodal},
it is unclear how to enforce the constraint for a candidate function to take
values on the manifold since their canonical forms are designed to act between
linear spaces.
%
%
%
To address this limitation, we consider in the following the Poincaré ball model and develop a neural architecture mapping the Euclidean space into the open unit ball. In particular,
let the element-wise hyperbolic tangent be defined as
\begin{gather}
h \colon \R^k \to \{\vx \in \R^k : \|\vx\|_\infty < 1\} \\
(x_1, \ldots, x_k) \mapsto (\tanh x_1, \ldots, \tanh x_k),
\end{gather}
which maps a linear space onto the open $\ell_\infty$ ball. Moreover, we define a ``squashing`` function
\begin{gather}
s \colon \{\vx \in \R^k : \|\vx\|_\infty < 1\} \to \{\vx \in \R^k : \|\vx\|_2 < 1 \} \\
  s(\vx) = \begin{cases} \vx \mapsto \vx \frac{\|\vx\|_\infty}{\|\vx\|_2}, & \text{if } \vx \neq \vec{0} \\
\vec{0}, & \text{if } \vx = \vec{0}
\end{cases}
\end{gather}
where $\vec{0}$ is the vector of all zeros. Since $\|\vx\|_{\infty} < \|\vx\|_2$, this function is continuous and maps the open $\ell_\infty$ ball into the open $\ell_2$ ball. And because both $s$ and $h$ are bijective continuous function with continuous inverse, the composition $s \circ h \colon \R^k \to  \{ \vx \in \R^k : \|\vx\|_2 < 1 \}$ is also a homeomorphism from $\R^k$ into the open ball $\ell_2$ and therefore also on the Poincaré model manifold. By composing $s \circ h$ with the neural network feature extractor $g_\theta$ we obtain a deep model that jointly learns features into a linear space and maps them to the hyperbolic manifold:
\begin{equation}
  f_{nng} = s \circ h \circ g_\theta \colon \R^d \to \manifold{P}^k.
\end{equation}
Note that the homeomorphism $ s \circ h  $ is sub-differentiable. Therefore learning the parameters $\theta$ of this model is akin to training a classical deep learning architecture with activation functions at the output layer corresponding to $s \circ h$. The key difference here lies in the loss used for training. In this setting, analogously to the task addressed by HSP, we replaced of the standard mean-squared error (Euclidean) loss with the squared geodesic distance between predictions and true labels.

\paragraph{Hyperbolic embeddings and manifold regression} In this work we propose to leverage the hyperbolic geometry to address machine learning tasks where hierarchical structures play a central role. In particular, we combine label embeddings approaches with hyperbolic regression to perform hierarchical classification. We do this by following a two step procedure: assuming a hierarcy $\mathcal{T}$, we consider an augmented $\mathcal{T}_{\X}$ where each example $x_{i}$ corresponds to a child to its associated class $c_i$ from the original $\mathcal{T}$. Then, we embed $\mathcal{T}_{\X}$ into the hyperboilic space using the procedure reviewed in \Cref{sec:hyper}. We compute similarity scores $\gamma(\cdot, \cdot)$ in the transitive closure of $\mathcal{T}_\X$, using either a Gaussian kernel on the features -- when both nodes have a corresponding representation available -- or otherwise employing the original $\gamma$. This allows us to incorporate information about feature similarities within the label embedding.

%% file: sections/applications.tex
We evaluate our proposed methods for hyperbolic manifold
regression on the following experiments:

\emph{Hierarchical Classification via Label Embeddings.} For this task, the goal
is to classify examples with a single label from a class hierarchy with tree
structure. We begin by computing label embeddings of the class hierarchy via hyperbolic representations. We then learn to regress
examples onto label embeddings and classify them using the nearest label in the
target space, i.e., by denoting $\vec{y}_c \in \Lor^{n}$ the embedding of class $c$ and taking $f:\R^{d} \to \Lor^{n}$.
\begin{equation}
  \widehat{c} = \argmin_{c \in \mathcal{C}} d(f(\vx), \vec{y}_c)
  \label{eq:class-nn}
\end{equation}

\emph{Taxonomy expansion.} For this task, the goal is to expand an existing
taxonomy based on feature information about new concepts. As for hierarchical
classification, we first embed the existing taxonomy in hyperbolic space and
then learn to regress onto the label embeddings. However, a key
difference is that a new example $c$ can
themselves act as the parent of another class $c^\prime$.


\paragraph{Models and training details}
For hierarchical classification, we compare to standard baselines such as
top-down classification with logistic regression (TD-LR) and hierarchical SVM
(HSVM). Furthermore, since both tasks can be regarded as regression problems
onto the Poincaré ball (which has a canonical embedding in $\R^k$) we also
compare to kernel regularized least squares regression (KRLS) and a neural
network with squared Euclidean loss (NN-E). In both cases, we constrain
predictions to remain within the Poincaré ball via the projection
\begin{equation*}
  \text{proj}(\vy) = \begin{cases}
  \vy / \|\vy\| - \varepsilon & \text{if }\|\vy\| \geq 1 \\
  \vy & \text{otherwise}
  \end{cases},
\end{equation*}
where \(\varepsilon\) is a small constant to ensure numerical stability,
equal to \(\varepsilon = 10^{-6}\). These regression baselines allows
us to evaluate the advantages of training manifold-valued models with squared
geodesic loss compared to standard methods that are agnostic of the underlying
geodesics.

\begin{table*}[t!]
    \small
    \caption{Hierarchical classification on benchmark datasets. We report micro-F1
        ($\mu$F1), macro-F1 (MF1), as well as the rank relative to all other models
        on a dataset, e.g., (1) for the the best performing model.}\label{tab:ns20}
    \begin{tabular}{llcccccccccc}
        \toprule
        & & \multicolumn{10}{c}{\bf Model - Performance (Relative Rank)} \\
        \cmidrule(r){3-12}
        & & \multicolumn{2}{c}{TD-LR} & \multicolumn{2}{c}{HSVM} &  \multicolumn{2}{c}{NN-E}  & \multicolumn{2}{c}{NN-G} & \multicolumn{2}{c}{HSP} \\
        \midrule
        \multirow{2}{*}{\bf News-20}
        & $\mu$F1 & 77.07 & (3) & \bf{80.79} & (1) & 63.91 & (5) & 72.67 & (4) & 80.28
        & (2) \\
        & MF1 & 77.94 & (3) & \bfseries{80.04} & (1) & 64.21 & (5) & 72.70 & (4) & 79.56 & (2) \\
        \cmidrule(r){1-2}\cmidrule(lr){3-4}\cmidrule(lr){5-6}\cmidrule(lr){7-8}\cmidrule(lr){9-10}
        \cmidrule(lr){11-12}
        \multirow{2}{*}{\bf Imclef07a}
        & $\mu$F1 & 73.86 & (3) & 74.98 & (2) & 65.49 & (5) & 67.49 & (4) & \bf{75.95} & (1) \\
        & MF1 & 36.03 & (3) & \bf{50.44} & (1) & 26.76 & (5) & 31.20 & (4) & 46.41 & (2)\\
        \cmidrule(r){1-2}\cmidrule(lr){3-4}\cmidrule(lr){5-6}\cmidrule(lr){7-8}\cmidrule(lr){9-10}
        \cmidrule(lr){11-12}
        \multirow{2}{*}{\bf Wipo}
        & $\mu$F1 & 36.85 & (2) & \bf{38.48} & (1) & 16.87 & (5) & 16.69 & (6) & 31.94 & (3)  \\
        & MF1 & 52.18 & (3) & 52.21 & (2) & 42.77 & (5) & 42.86 & (4) & \bf{52.41} &
        (1) \\
        \cmidrule(r){1-2}\cmidrule(lr){3-4}\cmidrule(lr){5-6}\cmidrule(lr){7-8}\cmidrule(lr){9-10}
        \cmidrule(lr){11-12}
        \multirow{2}{*}{\bf Diatoms}
        & $\mu$F1 & \textbf{54.01} & (1) & 48.97 & (3) & 9.25 & (5) & 11.31 & (4) &
        53.20
        & (2) \\
        & MF1       & 55.53 & (2) & 44.61 & (3) & 14.90 & (4) & 14.61 & (5) &
        \bf{62.10}
        & (1)  \\
        \midrule
        & Avg. Rank & \multicolumn{2}{r}{(2.5)} & \multicolumn{2}{r}{(1.75)} & \multicolumn{2}{r}{(4.88)} & \multicolumn{2}{r}{(4.38)} & \multicolumn{2}{r}{(1.75)}  \\
        \bottomrule
    \end{tabular}
\end{table*}

For kernel-based methods, we employ a Gaussian kernel selecting the  bandwidth
$\sigma \in [10^{-1}, 10^{2}]$ and
regularization parameter $\lambda \in [10^{-6}, 10^{-2}]$ via
cross-validation. Both parameter ranges are logarithmically spaced. For HSP
inference we use RSGD with batch size equal to $50$ and a maximum of $40000$
iterations. We stop the minimization if the the gradient Euclidean
norm is smaller than $10^{-5}$ (In most cases the inference stops before the
  $10000$ iteration). The learning rate for RSGD is chosen via
cross-validation on the interval $[10^{-5}, 10^{-1}]$. For the neural network
models (NN-G, NN-E) we use the same architecture for
$g_{\theta}$: each layer is a fully connected network
\begin{equation*}
  z^{\ell} = \psi(W_{\ell}z^{\ell - 1} + b_{\ell})
\end{equation*}
where ${\psi(x) = \max(0, x)}$ is a ReLU non-linearity and $\theta = \{W \in \R^{s/2 \times s}, b \in \R^{s/2}\}$, with $s$ the dimension of the previous layer (with the exception
of the first and last layer which must fit input and output dimensions). We use a
depth of 5 layers with intermediate dimensionalities $s \in (1024, 1024, 512,
  256, 128)$ for taxonomy expansion and $s \in (2048, 2048, 1024, 512,
  256)$ for hierarchical classification. We did not find significant improvements
with deeper architectures in performance. We train the deep models using mini-batch
stochastic gradient descent, with a scheduler until the model reaches convergence on the training loss.
For taxonomy expansion we also compare our algorithms with a hyperbolic
neural networks (HNN) as introduced by \cite{Ganea2018}. This architecture is trained with
Riemannian Stochastic Gradient Descent until convergence and has the same structure and the same
number of parameters of NN-G and NN-E. Because NN-G uses fully connected layers until the homeomorphic transformation, it can be trained with traditional optimizers
such as stochastic gradient descent or Adam~\citep{kingma2013auto}. In our
experiments, we observe that this can be an important advantage as these models
require typically one third of the training time compared to HNNs.

\subsection{Hierarchical classification}
For hierarchical classification, we are given a supervised training set
$\mathcal{D} = \{x_i, c_i\}_{i=1}^{m}$ where the class labels $c_i$ are organized in a
tree $\mathcal{T}$. We first embed the augmented hierarchy $\mathcal{T}_{x}$ as
discussed in \Cref{sec:method} and learn a regression function $ \hat{f} \colon \R^d \to \Lor^n$
using $\mathcal{D}^{e} =\{x_i, y_i\}_{i=1}^m$.
For a test point $\vx^\prime \in \R^d$, we first map it onto the target manifold
$\hat{y} = \hat{f}(\vx^{\prime})$ and then classify $\hat{y}$ according to
\Cref{eq:class-nn}.
For evaluation, we use various benchmark datasets for hierarchical classification\footnote{\url{https://sites.google.com/site/hrsvmproject/}},
and Newsgroups-20\footnote{\url{http://qwone.com/~jason/20Newsgroups/}}
for which we manually extract TF-IDF
features $x_i \in  \R^{10000}$ from the original documents. We compute an embedding for the augmented hierarchies
of each dataset. To make sure to obtain a good embedding, we perform parameter-tuning in order to attain mAP of at least $0.99$. We
then train HSP, NN-G and NN-E as described above
and measure classification performance in terms of $\mu$F1 and macroF1 scores.
As a baseline we also train Hierarchical SVM (HSVM)~\citep{vateekul2012top} and
Top-Down Logisitic Regression (TD-LR)~\citep{Naik2018}.

\Cref{tab:ns20} shows the results of our experiments. It can be
seen that the hyperbolic structured predictor achieves
results comparable to state-of-the-art on this task although we did not explicitly optimize the embedding
or training loss for hierarchical classification. We also observe that while NN-G outperforms NN-E, both algorithms perform significantly worse on Wipo and Diatoms datasets. Interestingly, these two datasets are significantly smaller compared to Newsgroup-20 and Imclef07a in terms of number of training points ($\sim1K$ Vs $\sim10K$ training samples). This seems to suggest that NN-G and NN-E models have a higher sample complexity.

\begin{figure*}[!tbp]
    \begin{subfigure}[b]{0.48\textwidth}
        \includegraphics[width=\textwidth]{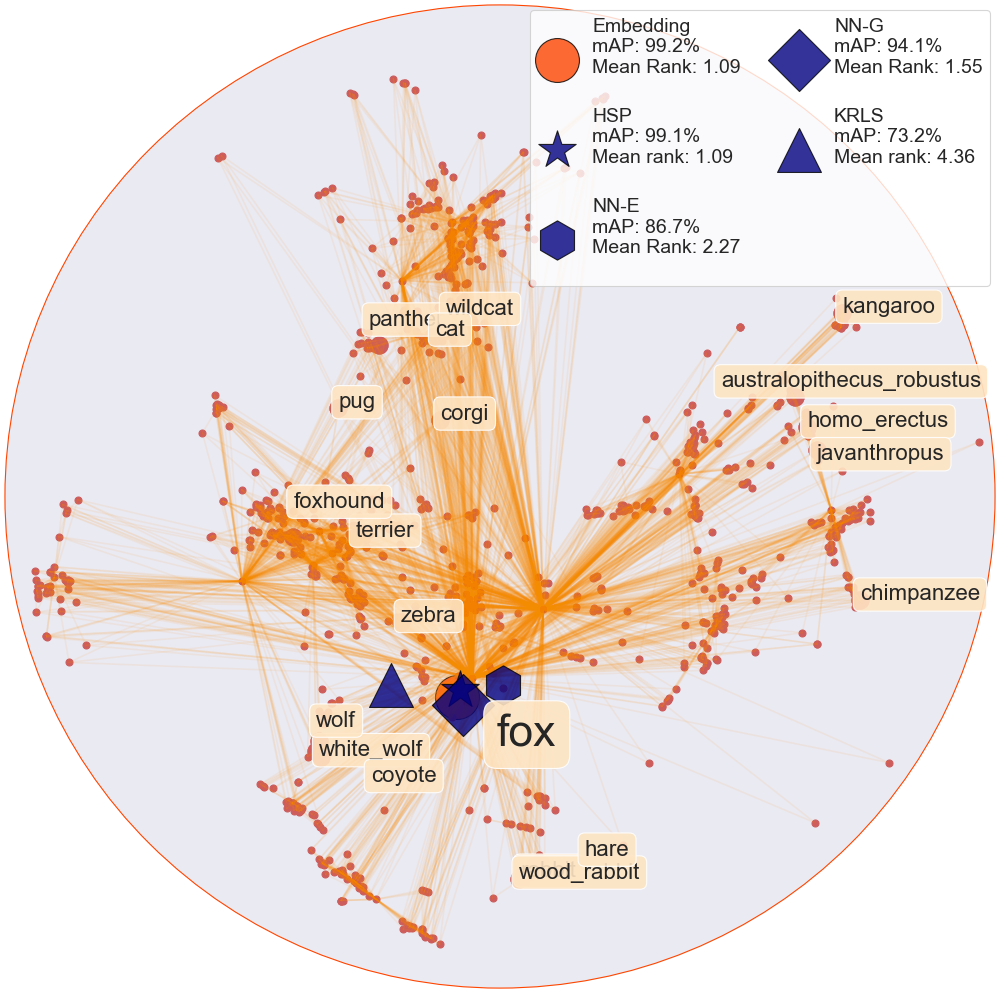}
        \caption{WordNet mammals embedding}
        \label{fig:f1}
    \end{subfigure}
    \hfill
    \begin{subfigure}[b]{0.48\textwidth}
        \includegraphics[width=\textwidth]{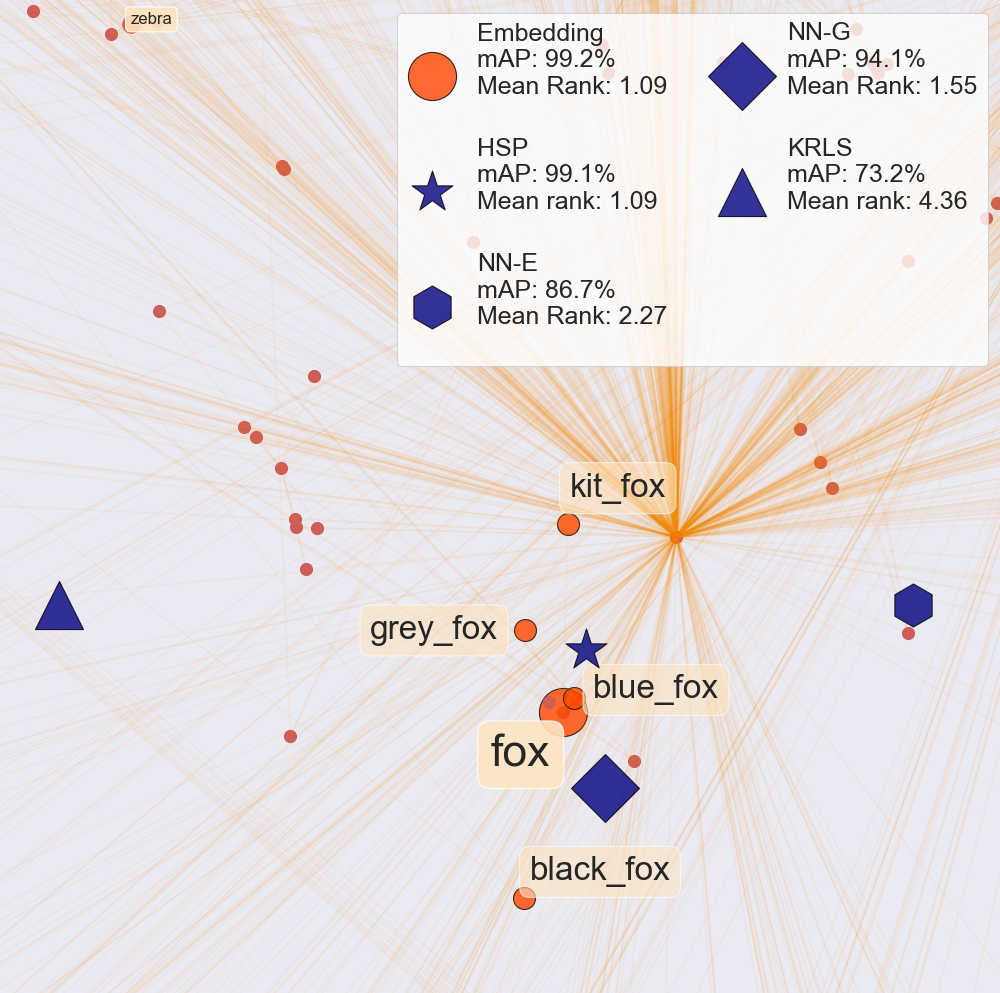}
        \caption{Close-up of predicted embedding for 'Fox'}
        \label{fig:f2}
    \end{subfigure}
    \caption{Overview and close-up of predicted positions for entity 'Fox'. Models that do not use the geometry of the hyperbolic manifold fail at positioning the entity, while the geodesic neural network and the hyperbolic structured predictor position the entity accordingly to its real neighbours.}
\end{figure*}

\subsection{Taxonomy expansion}
For taxonomy expansion, we assume a similar setting as for hierarchical
classification. We are given a dataset $\mathcal{D} = \{x_i, c_i\}_{i=1}^{m}$
where concepts $c_i$ are organized in a taxonomy $\mathcal{T}$ and for each
concept we have an additional feature representation $x_i$. Again, we first
embed the augmented hierarchy $\mathcal{T}_{x}$ as discussed in
\Cref{sec:method} and split it in train $\mathcal{D}_\text{train}^e$ and test set
$\mathcal{D}_\text{test}^e$. We vary the size of the test set, i.e., the number of
unknown concepts in $\mathcal{T}$ such that $|\mathcal{D}_\text{test}| \in \{5,10,
20, 30, 50\}$. Whenever necessary, we also create a validation set from
$\mathcal{D}_\text{train}$ for model selection with a $80:20$ ratio for model
selection. We then train all regression functions $ \hat{f} \colon \R^d \to
\Lor^n$ using $\mathcal{D}_\text{train}^e$ and predict embeddings for $\mathcal{D}_{\text{test}}$.
In contrast to hierarchical classification, the predicted points $\hat{\vy} =
f(\vx)$ can themselves act as parents of other points, i.e., they are full
members of the taxonomy $\mathcal{T}$. To assess the quality of the predictions
we use mean average prediction (mAP) as proposed by~\cite{Nickel2017a}. We
report mAP for the predicted points as well as for the points originally
embedded by the Lorentz embedding (Orig). This experiment is repeated $20$ times
for a given size of the test set, each time selecting a new training-test split.
In our experiments, we consider the following datasets:

\begin{table*}
    \small
    \caption{Mean average precision for taxonomy expansion on WordNet mammals
      and synthetic data}\label{tab:mammals}
    \begin{tabular}{lllllll}
      \toprule
      & &\multicolumn{5}{c}{\bf Number of new concepts} \\
      \cmidrule(l){3-7}
        & & 5             & 10            & 20            & 30            & 50            \\
        \midrule
      \multirow{6}{*}{\bf \shortstack{Wordnet\\Mammals}}
      & Orig & $0.86\pm0.06$ & $0.88\pm0.06$ & $0.87\pm0.03$ & $0.87\pm0.03$ & $0.88\pm0.02$ \\
        \cmidrule(l){2-7}
        & KRLS   & $0.54\pm0.14$ & $0.37\pm0.07$  & $0.26\pm0.04$ & $0.22\pm0.03$ & $0.15\pm0.02$ \\
        & NN-E   & $0.61\pm0.12$ & $0.47\pm0.08$   & $0.38\pm0.04$ & $0.31\pm0.03$ & $0.20\pm0.03$ \\
        & NN-G   & $0.79\pm0.08$ &$\bf0.74\pm0.06$ & $0.63\pm0.06$ & $\bf0.61\pm0.06$ & $\bf0.50\pm0.04$ \\
        & HNN    & $\bf0.82\pm0.05$& $0.73\pm0.05$    & $0.63\pm0.04$ & $0.57\pm0.05$  & $0.46\pm0.04$ \\
        & HSP    & $0.72\pm0.10$ & $0.69\pm0.07$    & $\bf0.69\pm0.07$ & $0.58\pm0.09$ & $\bf0.50\pm0.06$ \\
        \midrule
        \multirow{6}{*}{\bf \shortstack{Synthetic\\Small}} & Orig     & $0.94\pm0.03$ & $0.93\pm0.03$ & $0.94\pm0.02$ & $0.94\pm0.02$ & $0.94\pm0.01$ \\
        \cmidrule(l){2-7}
        & KRLS     & $0.63\pm0.16$  & $0.51\pm0.12$     & $0.36\pm0.06$     & $0.27\pm0.03$ & $0.21\pm0.02$ \\
        & NN-E     & $0.76\pm0.07$  & $0.72\pm0.09$     & $0.63\pm0.09$     & $0.56\pm0.09$ & $0.45\pm0.08$ \\
        & NN-G     & $0.80\pm0.07$  & $0.73\pm0.06$     & $0.61\pm0.06$     & $0.55\pm0.05$ & $0.45\pm0.04$ \\
        & HNN   & $\bf0.82\pm0.01$ & $0.71\pm0.07$ & $0.60\pm0.05$ & $0.51\pm0.04$ & $0.41\pm0.04$\\
        & HSP      & $\bf0.82\pm0.08$ & $\bf0.76\pm0.07$ & $\bf0.66\pm0.05$ & $\bf0.60\pm0.04$ & $\bf0.50\pm0.03$ \\
        \midrule
        \multirow{6}{*}{\bf \shortstack{Synthetic\\ Large}} & Orig     & $0.81\pm0.06$ & $0.79\pm0.05$ & $0.80\pm0.03$ & $0.80\pm0.02$ & $0.80\pm0.01$ \\
        \cmidrule(l){2-7}
        & KRLS     & $0.30\pm0.05$  & $0.20\pm0.02$     & $0.13\pm0.01$     & $0.09\pm0.01$     & $0.07\pm0.00$ \\
        & NN-E     & $0.69\pm0.09$  & $0.68\pm0.09$     & $0.64\pm0.05$     & $0.61\pm0.05$     & $0.59\pm0.05$ \\
        & NN-G     & $0.77\pm0.07$ & $0.72\pm0.07$ & $0.71\pm0.04$ & $\bf0.69\pm0.04$   & $\bf0.65\pm0.03$ \\
        & HNN   & $\bf0.83\pm0.7$ & $\bf0.79\pm0.4$ & $\bf0.72\pm0.06$ & $0.64\pm0.06$ & $0.63\pm0.02$\\
        & HSP      & $0.76\pm0.09$  & $0.70\pm0.07$     & $0.69\pm0.04$     & $0.67\pm0.05$     & $0.63\pm0.03$ \\
        \bottomrule
\end{tabular}
\vspace{-0.2cm}
\end{table*}

{\em WordNet Mammals.} For WordNet Mammals, the goal is to expand an existing
taxonomy by predicting concept embeddings from text. For this purpose, we take
the mammals hierarchy of WordNet and retrieve for each node its corresponding
Wikipedia page. If a page is missing, we remove the corresponding node and if a
page has multiple candidates we disambiguate manually. The transitive closure of
$\mathcal{T}$ has $1036$ nodes and $11222$ edges. Next, we pre-process the
retrieved Wikipedia descriptions by removing all non alphabetical characters,
tokenizing words and removing stopwords using NLTK~\citep{Loper2002}. Finally, we
associate to each concept $c_{i} \in \mathcal{T}$ the TF-IDF vector of its
Wikipedia description as feature representation $x_{i} \in \R^{10000}$
computed using Scikit-learn~\citep{Pedregosa2011}. We then embed $\mathcal{T}$
following \cref{sec:hyper} and
obtain an embedding with mAP $0.86$ and mean rank $4.74$. This dataset is
particularly difficult given the way features were collected: Wikipedia pages
have a high variance in quality and amount of content, while some pages are
detailed and rich in information other barely contain a full sentence.

{\em Synthetic datasets.}
To better control for noise in the feature representations, we also generate datasets based on synthetic random trees, i.e., a smaller tree with $226$ nodes and $1228$ edges and a larger tree with $2455$ nodes and $30829$ edges after transitive closure. For each node we take as feature vector the corresponding row of the adjacency matrix of the transitive closure of the tree. We project these rows on the first $d$ principal components of the adjacency matrix, where $d = 50$ for the small tree and $d = 500$ for the big tree. We then embed the nodes of the graph in $\Lor^{5}$ using both the tree structure and similarity scores computed using the vector features. The similarity is computed by a Gaussian kernel with $\sigma$ equal to the average tenth nearest neighbour of the dataset.


\emph{Results} We provide the results of our evaluation for different sizes on
$\mathcal{D}_\text{test}^e$ in \cref{tab:mammals}. It can be seen that all hyperbolic-based methods can successfully predict the embeddings of unknown concepts when the
test set is small. The performance degrades as the size of the test set
increases, since it becomes harder to leverage the original structure of the graph.
While all methods are affected by this trend, we note that algorithms using the
geodesic loss tend to perform better than those working in the linear space.
This suggest that taking into account the local geometry of the embedding is
indeed beneficial in estimating the relative position of novel points in the
space. 

We conclude by noting that all hyperbolic-based methods have comparable performance across the three settings. However, we point out that HSP and NN-G offer significant practical advantages over HNN: in all our experiments they were faster to train and in general more amenable do model design. In particular, since HSP is based on a kernel method, it has relatively fewer hyperparameters and requires only solving a linear system at training time. NN-G consists of a standard neural architecture with the homeomorphism activation function introduced in \cref{sec:method} and trained with the geodesic loss. This allows one to leverage all current packages available to train neural networks, significantly reducing both modeling and training times.




%% file: sections/conclusions.tex
In this paper, we showed how to recast supervised problems with hierarchical structure as manifold-valued regressions in the hyperbolic manifold. We then proposed two algorithms for learning manifold-valued functions mapping from Euclidean to hyperbolic space: a non-parametric kernel-based
method for which we also proved generalization bounds and a parametric
deep-learning model that is informed by the geodesics of the output space.
The latter makes possible to leverage traditional neural network layers for regression on hyperbolic space without resorting to hyperbolic layers, thus requiring a smaller training time.
We evaluated both methods empirically on the task of hierarchical classification
and showed that hyperbolic structured prediction shows strong
generalization performance.
We also showed that hyperbolic manifold regression enables new applications in
supervised learning. By exploiting the continuous representation of hierarchies in hyperbolic space
we were able to place unknown concepts in the embedding of a taxonomy using manifold regression.
Moreover, by comparing to hyperbolic neural networks we showed that for this application, the key step is
leveraging the geodesic of the manifold.
In this work, we have aimed at developing a foundation for regressing onto
hyperbolic representations. In future work, we plan to exploit this framework in
dedicated methods for hierarchical machine learning and extending the applications to manifold product spaces.
